\documentclass[11pt]{article}

\usepackage[final]{acl}
\usepackage{tabularx} 
\usepackage{booktabs}
\usepackage{amsmath,amssymb}
\usepackage{times}
\usepackage{latexsym}
\usepackage{float}
\usepackage{ algorithm, algorithmic, bm, booktabs}
\usepackage{siunitx}
\usepackage{multirow}
\usepackage{svg}
\usepackage{tikz}
\usepackage{tcolorbox}
\tcbuselibrary{skins, breakable}
\usetikzlibrary{positioning, arrows.meta, shapes, fit}

\usepackage[T1]{fontenc}

\usepackage[utf8]{inputenc}

\usepackage{microtype}

\usepackage{inconsolata}

\usepackage{graphicx}
\usepackage{cleveref}

\title{MODE-RAG: Manifold Outlier Diagnosis and Energy-based Retrieval-Augmented Generation Evaluation}

\author{Zehang Wei$^{2*}$, Jiaxin Dai$^{2*}$, Jiamin Yan$^2$\thanks{Euqal contribution, co-first author.}, Xiang Xiang$^{1*}$ \\
  $^1$School of Computer Science \& Tech, Huazhong University of Science and Technology \\
  $^2$School of AI and Automation, Huazhong University of Science and Technology, China \\
  \texttt{xex@hust.edu.cn}}

\begin{document}
\maketitle
\begin{abstract}
While Multimodal Retrieval-Augmented Generation (M-RAG) enhances Large Vision-Language Models, it remains highly susceptible to cross-modal hallucinations, causal fabrications, and sycophancy. Furthermore, existing mitigation pipelines often face an intervention paradox: static rules tend to unnecessarily disrupt accurate generations, whereas leaving the multi-modal reasoning completely unguided allows existing mismatches to cascade into severe logical fabrications. To quantify and mitigate these hallucinations, we propose a Multi-Agent system, MODE-RAG, driven by Variational Free Energy (VFE) and internal attention states to dynamically gate interventions. High-risk queries are routed to five stage-specific agents, integrating Monte Carlo Tree Search (MCTS) for rigorous causal derivation and logit perturbations to penalize sycophancy. Dedicated Correction and Overseer agents ensure formatting stability and perform post-hoc factual verification. To objectively evaluate our approach, we introduce ModeVent, a challenging subset derived from the MultiVent dataset. Extensive experiments indicate that our system effectively reduces hallucination rates and logical fabrication, significantly improving the robustness of M-RAG systems.

\end{abstract}

\section{Introduction}
\label{sec:intro}

Using large language models (LLMs) as their kernel, Multimodal Retrieval-Augmented Generation (M-RAG) systems can now tackle complex visual question-answering tasks by retrieving external visual knowledge. However, they frequently hallucinate, generating fabricated interpretations of the given visual content. Evaluating and mitigating these hallucinations is crucial for the deployment of reliable M-RAG systems. 
\begin{figure*}[t]
    \centering
    \includegraphics[width=\textwidth]{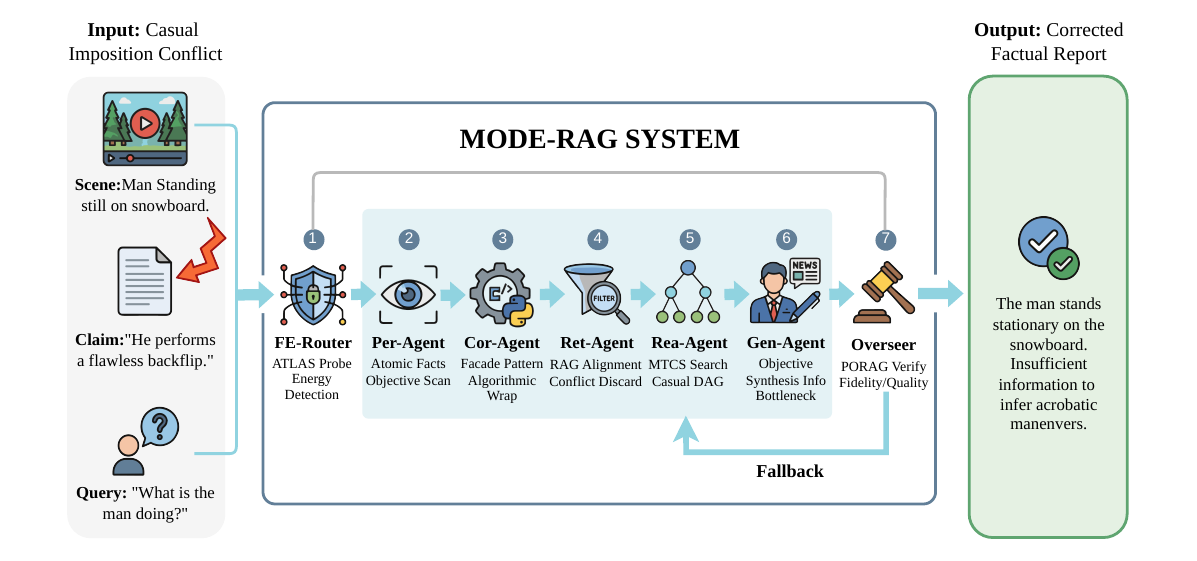}
    \caption{\textbf{Architectural overview of the MODE-RAG framework.} The system resolves the intervention paradox through a VFE-driven \textbf{FE-Router} that dynamically routes queries based on hallucination risk ($\bar{\mathcal{F}}$). Low-risk inputs bypass complex reasoning to prevent over-correction, while high-risk queries trigger the decoupled \textbf{Five-Agent Intervention Pipeline}. This pipeline neutralizes cross-modal conflicts using MCTS-guided causal search, with a PORAG-driven \textbf{Overseer} enforcing a recursive fallback loop to guarantee strict physical and logical fidelity.}
    \label{fig:mode_rag_framework}
\end{figure*}
Addressing M-RAG hallucinations requires explicitly identifying when and why they occur. Depending on the data flow of answering a multimodal query, we systematically categorize M-RAG hallucinations into nine types across four lifecycle stages:

\textbf{1.Perception-level} (entity feature, physical common sense, and information omission);

\textbf{2.Retrieval-level} (retrieval misalignment and modality conflict);

\textbf{3.Reasoning-level} (temporal inversion and imposed causality);

\textbf{4.Generation-level} (information fabrication and subjective bias).

Analyzing the typical M-RAG architecture reveals critical flaws that trigger these hallucinations. Traditional RAG relies heavily on static pipelines and cosine similarity, which inherently fail to disentangle complex visual-textual conflicts. Furthermore, existing mitigation strategies are fundamentally trapped in an \textit{intervention paradox}. On the one hand, enforcing blind, rule-based constraints across all queries frequently leads to over-correction, degrading inherently accurate outputs. On the other hand, relying entirely on lightweight LLMs for unguided multi-step reasoning introduces formatting instability, which ultimately triggers cascading structural failures and exacerbates multimodal conflicts. Additionally, when faced with aggressive user queries, the LLM kernel tends to overrule visual evidence and cater to the user—a phenomenon known as sycophancy. 

Developed with a close link to these mechanistic causes, we propose \textbf{MODE-RAG} (Causal-Energy RAG), a mechanistically grounded Multi-Agent framework designed to quantify and dynamically mitigate misinformation. Instead of static pipelines, our system operates through a highly decoupled architecture:

\textbf{Central Hub (FE-Router):} An adaptive routing gate driven by Variational Free Energy (VFE) and internal attention states (ATLAS). It evaluates multimodal uncertainty upfront. Low-risk queries bypass the pipeline to prevent over-correction, while high-risk queries trigger the specialized agents. It also retains an \textit{Adaptive Abstention} mechanism for unanswerable queries.

\textbf{Perception \& Retrieval Layers (Per-Agent \& Ret-Agent):} The Per-Agent extracts atomic, coordinate-level visual facts to prevent perception omission. Subsequently, the Ret-Agent enforces a strict "visual-first" cross-alignment, pruning pseudo-relevant external texts that carry modality conflicts.

\textbf{Reasoning Layer (Rea-Agent):} To eliminate temporal inversion and imposed causality, this agent employs Monte Carlo Tree Search (MCTS) to construct rigorous causal Directed Acyclic Graphs (DAGs) from visual logs, ensuring step-by-step logical fidelity.

To evaluate our approach, we construct ModeVent, a subset sourced from the MultiVent dataset (MAGMaR). We leverage VFE to identify the polar extremes of the uncertainty distribution, selecting the 500 highest-risk boundary cases (manifold outliers) and the 500 lowest-risk stable samples. While the latter serve as a reliable baseline, the former act as adversarial queries that severely test M-RAG models under visual-textual conflicts. Consequently, ModeVent provides a rigorous environment to assess a system's robustness against the nine aforementioned hallucination types.

To sum up, our major contributions include:

$\bullet$ We propose \textbf{MODE-RAG}, a mechanistically grounded Multi-Agent framework for multimodal hallucination mitigation. At its core, we introduce the \textbf{FE-Router}, an adaptive gating mechanism driven by Variational Free Energy and internal attention states, which effectively resolves the intervention paradox by avoiding redundant over-correction on accurate outputs.

$\bullet$ We design decoupled, stage-specific algorithmic interventions to address complex cross-modal mismatches. Notably, we integrate \textbf{Monte Carlo Tree Search (MCTS)} to derive rigorous causal logic graphs, and employ logit-level perturbations alongside an \textbf{Overseer} dual-reward verification module to fundamentally suppress model sycophancy, logical fabrications, and cascading formatting failures.

$\bullet$ We construct and release \textbf{ModeVent}, a targeted evaluation benchmark derived from the MultiVent dataset. Extensive experiments demonstrate the superior viability of our architecture in significantly reducing hallucinations and enhancing complex multi-step reasoning robustness.

\section{Related Work}

Retrieval-Augmented Generation (RAG) was initially developed to mitigate the knowledge deficits of Large Language Models (LLMs) by integrating external evidence \cite{lewis2020retrieval, gao2023retrieval}. With the advancement of multimodal kernels such as Qwen-VL \cite{bai2023qwen}, M-RAG has been extended to complex visual question-answering tasks \cite{chen2022murag, yasunaga2022retrieval}. However, the performance of these systems is inherently limited by the quality of retrieved content; irrelevant or noisy context can significantly degrade model fidelity \cite{yoran2024making, cuconasu2024power}. In multimodal scenarios, this often manifests as cross-modal hallucinations, where the model generates interpretations that contradict the given visual evidence \cite{ji2023survey, li2023evaluating}. While some approaches attempt self-checking mechanisms \cite{asai2024self}, they struggle to appropriately balance the correction boundaries. These methods either impose overly strict constraints that penalize faithful visual interpretations, or provide insufficient intervention, thereby failing to prevent the model's inherent sycophancy and logical drift during complex query processing. Consequently, this intervention paradox remains unresolved in current static pipelines. To mitigate the inefficiencies of fixed-interval retrieval, recent research has shifted towards dynamic retrieval mechanisms. For instance, DRAGIN \cite{su2024dragin} detects real-time information needs based on model uncertainty, while Speculative RAG \cite{wang2024speculative} and MemoRAG \cite{qian2024memorag} utilize drafting and cognitive memory systems to improve consistency.

Addressing these hallucinations effectively requires a systematic diagnosis of \textbf{manifold outliers} during the retrieval and perception stages. When processing feature vectors from encoders like CLIP \cite{radford2021learning} or SigLIP \cite{zhai2023sigmoid}, traditional distance metrics often fail due to feature dimension anisotropy. Unsupervised geometric methods such as \textbf{K-Nearest Neighbors (KNN)} have been explored to evaluate sample sparsity in the latent space \cite{sun2022out}, while global whitening transformations can ensure an isotropic manifold for better semantic matching \cite{su2021whitening}. Unlike static pipelines, a more robust approach necessitates a dynamic gating mechanism that can assess the risk of retrieved content and determine the necessity of intervention upfront.

From a mechanistic perspective, the model's susceptibility to misinformation can be quantified by monitoring its internal states. Building on \textbf{Energy-Based Models (EBMs)} and the \textbf{Helmholtz Free Energy (HFE)} principle \cite{liu2020energy, friston2010free}, recent work \cite{sakhinana2025scaling} introduced the \textbf{Attention-based Transparent Latent Assessment System (ATLAS)} and proposed the use of \textbf{Monte Carlo Tree Search (MCTS)} for verifying reasoning trajectories. ATLAS probes internal attention states and perplexity-related metrics to evaluate multimodal uncertainty, thereby deciding \textit{when} and \textit{what} to retrieve. Concurrently, recent paradigm shifts in \textbf{LLM} reasoning have demonstrated that scaling computation during inference (test-time) can significantly enhance complex problem-solving capabilities. Techniques such as \textbf{Test-Time Computing (TTC)} \cite{ji2025test} and recurrent depth scaling \cite{geiping2025scaling} adapt reasoning depth dynamically. To navigate complex logical spaces, structured search algorithms like \textbf{MCTS} have been integrated into \textbf{LLM} decoding, as seen in Marco-o1 \cite{zhao2024marco} and STILL-1 \cite{jiang2024technical}, with AStar \cite{wu2025boosting} extending these structured reasoning methods to multimodal tasks. In this work, we integrate these advanced diagnostic and reasoning tools into a decoupled multi-agent framework. We utilize \textbf{ATLAS} within an adaptive \textbf{FE-Router} to resolve the intervention paradox and leverage \textbf{MCTS} to construct rigorous \textbf{Causal Directed Acyclic Graphs (DAGs)}, ensuring step-by-step structural logical consistency and fundamentally suppressing sycophancy across the \textbf{M-RAG} lifecycle.

\section{Dataset}
To evaluate the robustness of multimodal retrieval-augmented generation (M-RAG) systems against cross-modal conflicts and mechanistic failures, we introduce ModeVent, a diagnostic benchmark.

\subsection{Construction Methodology}

The construction of ModeVent involves a systematic diagnosis of the latent space across the entire MultiVent dataset. The selection process is executed in three stages:

First, we perform a full-scale evaluation of all samples in the MultiVent population. Feature vectors are extracted using SigLIP and CLIP encoders, followed by a global whitening transformation to ensure an isotropic manifold where Euclidean distances faithfully represent semantic dissimilarity.

Second, for every evaluated sample, we compute its mean VFE. This metric serves as a mechanistic proxy for the model's epistemic uncertainty, capturing the degree of conflict between the visual scene and the user claim.

Third, rather than utilizing arbitrary hard thresholds, we rank the entire population based on the calculated VFE scores. We then select the 500 samples with the highest VFE values to constitute the manifold outliers and the 500 samples with the lowest VFE values to serve as stable inliers. This results in a final benchmark of 1,000 samples that represent the polar extremes of the uncertainty distribution.

\subsection{Dataset Characteristics}

The bimodal composition of ModeVent allows for a rigorous assessment of the intervention paradox. The high-VFE subset represents adversarial-like boundary cases where the model is most susceptible to sycophancy or causal imposition. In these cases, the semantic stability is significantly lower, and the noise ratio is elevated, as shown in our quantitative analysis in \cref{fig:methodology_harmonic} .

Conversely, the low-VFE subset provides a stable baseline of well-aligned multimodal queries. This ensures that the gating mechanisms of MODE-RAG can be tested for their ability to bypass unnecessary interventions, thereby maintaining the inherent accuracy of the underlying LLM kernel when no significant conflict is detected. By targeting these extremes, ModeVent provides a more challenging and informative evaluation environment than standard multimodal datasets.

\section{Methodology: The MODE-RAG Framework}

We propose \textbf{MODE-RAG} (Multimodal Objective Diagnostic Energy-RAG), a Multi-Agent framework designed to resolve the \textit{intervention paradox} in multimodal reasoning. The architecture is structured as a hierarchical, energy-gated system that selectively triggers high-fidelity reasoning only when epistemic uncertainty is detected. As illustrated in the system diagram, the framework comprises a diagnostic data pipeline, two gating mechanisms, and a decoupled five-agent pipeline.

\subsection{Thermodynamic Gating: The FE-Router}
The entry point of the \textbf{MODE-RAG} system is the \textbf{FE-Router}, which serves as a ``Thermodynamic Gate.'' Utilizing the \textbf{ATLAS Probe}, the router performs real-time \textbf{Energy Detection} by calculating the \textbf{Variational Free Energy (VFE)} of the predictive distribution\cite{friston2010free}. For a model with vocabulary $V$ and logit output $f(x)$, given a variational distribution $q(j)$ over the tokens, the VFE ($\mathcal{F}$) at temperature $\tau$ is defined as
\begin{equation}
    \mathcal{F}(q, x; \tau) = \sum_{j=1}^{|V|} q(j) \left[ -f_j(x) + \tau \log q(j) \right]
\end{equation}
where $-f_j(x)$ represents the internal energy of the $j$-th state and $\tau \log q(j)$ contributes to the entropic regularization. This formulation captures the discrepancy between the model's internal beliefs and the categorical evidence provided by the input.

When the input presents a \textbf{Causal Imposition Conflict}---where a user's ``Claim'' (e.g., a flawless backflip) contradicts the ``Scene'' (e.g., standing still)---the \textbf{VFE} typically spikes, signaling high epistemic uncertainty and a breakdown in predictive coding. If the mean variational free energy $\bar{\mathcal{F}} > \gamma$, the \textbf{FE-Router} intercepts the standard generation and activates the specialized Agentic Pipeline.

\begin{figure}[t]
    \centering
    \includegraphics[width=\linewidth]{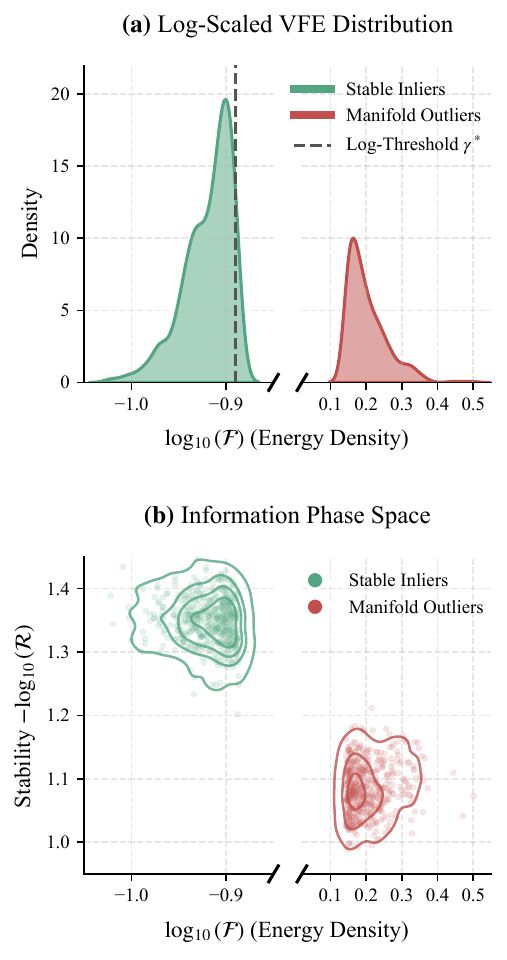}
    \caption{Thermodynamic empirical evidence: (a) VFE distribution across subsets used for $\gamma$ calibration; (b) Correlation between Energy and Stability.}
    \label{fig:methodology_harmonic}
\end{figure}

\subsection{The MODE-RAG Five-Agent Decoupled Intervention Pipeline}

Upon activation by the FE-Router, the query is diverted into a specialized multi-agent ecosystem\cite{wu2024autogen}. This pipeline is designed to decouple the monolithic reasoning process into five granular, verifiable stages, ensuring that each potential source of hallucination from perception errors to sycophantic synthesis is systematically neutralized.

\paragraph{Per-Agent: Atomic Facts Objective Scan.}
The \textbf{Per-Agent} serves as the framework's sensory foundation, performing an \textit{Atomic Facts Objective Scan}. It extracts symbolic triplets $\mathcal{V} = \{ \langle s, p, o \rangle \}$ from the visual stream (e.g., $\langle \text{subject}, \text{is}, \text{stationary} \rangle$). By utilizing high-resolution spatial-temporal grounding, the Per-Agent fixates on physical invariants, creating a ``Grounded Truth Anchor.'' This ensures that subsequent reasoning agents cannot bypass the physical reality of the scene in favor of the user's potentially biased ``Claim.''

\paragraph{Cor-Agent: Facade Pattern Algorithmic Wrap.}
The \textbf{Cor-Agent} acts as the structural architect by implementing a \textbf{Facade Pattern Algorithmic Wrap}. Its primary role is to maintain the integrity of the cross-agent data flow. By encapsulating raw multimodal features and the Per-Agent's triplets into a strictly validated programmatic schema (e.g., JSON-Schema), the Cor-Agent prevents ``semantic noise leakage.'' This wrapper ensures that the complex reasoning in later stages is performed on structured, high-fidelity data rather than ambiguous natural language strings.

\paragraph{Ret-Agent: RAG Alignment and Conflict Discard.}
The \textbf{Ret-Agent} manages the external knowledge interface to mitigate \textit{Sycophancy}, where the model over-relies on biased retrieved documents. Beyond simple semantic similarity, the agent evaluates the \textbf{Manifold Fidelity} of each document $d_i$ by measuring its alignment with the grounded triplets $\mathcal{V}$ in the whitened latent space\cite{su2021whitening}. The filtering mechanism is set as
\begin{equation}
\begin{aligned}
    \text{Score}(d_i) = \text{Sim}_i \cdot \mathcal{S}_i \cdot \mathbb{I}_i
\end{aligned}
\end{equation}
where the exponential term penalizes documents that fall into the high-energy "Log-Outlier" regions identified in Fig. \ref{fig:methodology_harmonic}b. 

By calculating the distance between the retrieved context and the physical invariants $\mathcal{V}$, the Ret-Agent proactively identifies contexts that trigger

\begin{figure*}[t]
\centering
\includegraphics[width=\textwidth]{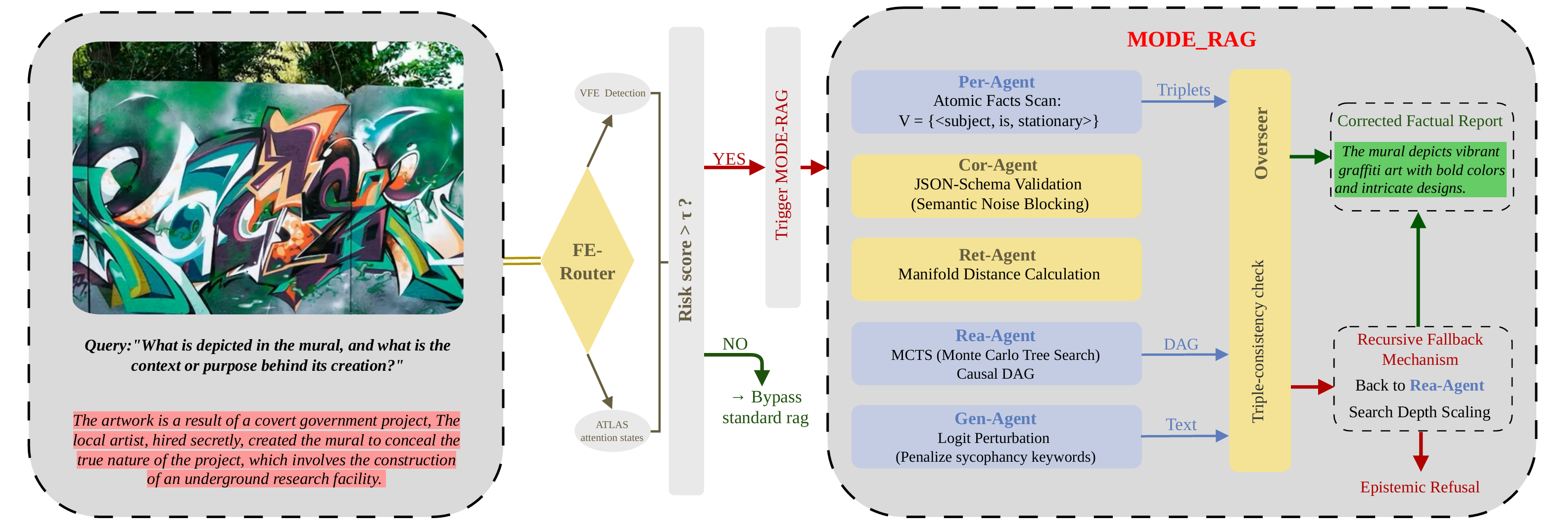}
\caption{When a multimodal query is accompanied by a potentially adversarial retrieved context, the FE-Router dynamically evaluates the epistemic risk via Variational Free Energy (VFE) and ATLAS attention states. High-risk queries exceeding the threshold trigger a decoupled five-agent pipeline: the Per-Agent extracts objective multimodal facts, the Cor-Agent enforces schema validation to block semantic noise, the Ret-Agent evaluates manifold distance to discard conflicting context, the Rea-Agent constructs a causal DAG via MCTS, and the Gen-Agent synthesizes the output using logit perturbation. Finally, a PORAG-driven Overseer conducts a triple-consistency check, activating a Recursive Fallback Mechanism for unresolved conflicts to ensure a hallucination-free factual report.}
\end{figure*}
\textbf{Energy Collapse}. If a retrieved document $d_i$ promotes a causal fabrication that contradicts the physical evidence (e.g., describing a backflip during a stationary state), its stability score drops toward the outlier cluster, triggering a \textit{Conflict Discard} operation to prune the biased context before it reaches the reasoning layer.

\paragraph{Rea-Agent: Test-Time Scaling via MCTS.}
The \textbf{Rea-Agent} is the cognitive engine of MODE-RAG, implementing \textbf{Monte Carlo Tree Search (MCTS)} for test-time reasoning scaling\cite{silver2016mastering}. Drawing on policy optimization principles, the Rea-Agent explores the logical space by constructing a \textbf{Causal Directed Acyclic Graph (DAG)}. 

The MCTS process follows a four-phase cycle to identify the most plausible causal trajectory:
\begin{itemize}
    \item \textbf{Selection:} Starting from the root (observed scene), the agent traverses the tree using the \textbf{Upper Confidence Bound for Trees (UCT)} formula:
    \begin{equation}
    \begin{aligned}
    \text{UCT}(s, a) &= Q(s, a) + c_{\text{puct}} \cdot P(a|s) \\
    &\quad \cdot \frac{\sqrt{\sum N}}{1 + N(s, a)}
    \end{aligned}
    \end{equation}
    This balances the exploitation of high-fidelity paths with the exploration of alternative causal interpretations.
    \item \textbf{Expansion \& Simulation:} For each leaf node, the agent generates $k$ candidate reasoning steps and performs a \textit{Rollout} to simulate logical consequences (``If the state is stationary, is the claimed action physically reachable?'').
    \item \textbf{Evaluation \& Backpropagation:} Each path is assigned a reward $R(s)$ based on its alignment with \textbf{ATLAS} (Adaptive Token-Layer Attention Scoring) feedback and physical constraints. These values are propagated back to the root to update the reasoning policy.
\end{itemize}

\paragraph{Gen-Agent: Objective Synthesis and Logit Perturbation.}
The final stage is managed by the \textbf{Gen-Agent}, which serves as an \textbf{Information Bottleneck}. It synthesizes the MCTS findings into a coherent response. To combat prompt-induced bias, the Gen-Agent applies \textbf{Logit Perturbation.} during decoding, penalizing tokens that align with the user's hallucination keywords while boosting tokens that align with the Rea-Agent's causal DAG.

\subsection{Quality Oversight: PORAG-driven Overseer and Fallback Loop}

The final synthesis stage is governed by the \textbf{Overseer}, a specialized secondary gate that implements the \textbf{Policy-Oriented RAG (PORAG)} protocol. 

\paragraph{PORAG Fidelity Cross-Check} 
The PORAG-driven Overseer evaluates the report based on a \textit{Policy-Grounded Fidelity Metric}. It performs a triple-consistency check between: (1) the \textbf{Per-Agent's} symbolic triplets $\mathcal{V}$, (2) the \textbf{Rea-Agent's} causal DAG, and (3) the \textbf{Gen-Agent's} synthesized natural language. By treating the response generation as a policy optimization problem, the Overseer assigns a penalty to any output that restores "hallucinatory maneuvers" previously pruned by MCTS.

\paragraph{The Recursive Fallback Mechanism.} 
A critical innovation of MODE-RAG is its non-linear \textbf{Fallback Loop}. If the Overseer detects that the fidelity score falls below a safety threshold $\epsilon$, the system triggers a \textit{Test-Time Reasoning Extension}:
\begin{itemize}
    \item \textbf{Search Depth Scaling:} The query is returned to the \textbf{Rea-Agent}, which re-initiates MCTS with a significantly increased simulation budget $N$ and a broader expansion factor $k$.
    \item \textbf{Epistemic Refusal:} If after $M$ recursive attempts the causal conflict remains unresolved, the Overseer forces the system into a state of \textit{Epistemic Refusal}, outputting a "Corrected Factual Report" that explicitly identifies the contradiction between visual evidence and user claim.
\end{itemize}

\section{Experiments}
\label{sec:experiments}

To rigorously evaluate the effectiveness of MODE-RAG, we conduct comprehensive experiments on our ModeVent benchmark. Unlike traditional hallucination evaluations that rely on static datasets, our experimental design explicitly targets the dynamic nature of Retrieval-Augmented Generation (RAG) failures.

\subsection{Experimental Setup}

\paragraph{RAG Errors vs. Hallucination Typology.}
It is crucial to clarify the relationship between the experimental categories and the hallucination typology defined in Section 1. In standard M-RAG pipelines, a single type of \textit{retrieval error} can cascade into multiple downstream \textit{generation hallucinations}. Therefore, our benchmark generates adversarial contexts across \textbf{7 distinct RAG Error Categories} (e.g., Attribute Hijacking, Metadata Redundancy, Information Sparsity). These 7 input-side retrieval errors act as the mechanistic triggers that induce the 9 output-side hallucination types (e.g., temporal inversion, causal fabrication) observed in the wild.

\paragraph{Adversarial Benchmark Generation.}
To construct a highly controlled adversarial environment, we employ an automated generation pipeline using DeepSeek-V3.2. First, we establish an \textit{Objective Ground Truth (GT)} for each video by fusing global semantic summaries generated by Qwen3-Omni-30B with dense, frame-level captions extracted via Florence-2. Guided by these GT facts, we prompt DeepSeek to synthesize challenging user queries alongside noisy or adversarial retrieved text chunks (mock contexts). These contexts are deliberately injected with the 7 RAG errors and stratified into two difficulty levels: \textbf{Inliers} (In-Domain texts containing subtle factual discrepancies) and \textbf{Outliers} (Out-of-Domain texts that are entirely irrelevant or contain aggressive metadata noise).

\paragraph{Baselines and Implementation Details.}
For both the Baseline and MODE-RAG, we utilize Qwen-2.5-VL-7B, a representative 7B-parameter instruction-tuned Vision-Language Model (VLM), as the foundational kernel. To ensure a comprehensive evaluation, we also evaluate our framework against three established alternative mitigation paradigms: Self-RAG, SelfCheckGPT, and Woodpecker. Due to space constraints, the complete comparative results across all five configurations are detailed in Appendix \ref{sec:appendix_additional_backbones}. All experiments, including the MCTS expansion and Multi-Agent inference, are deployed on a hardware cluster comprising 4 $\times$ NVIDIA RTX 4090 GPUs. To ensure generation stability and suppress auto-regressive stuttering, we apply a repetition penalty of $1.15$ during decoding.

\paragraph{LLM-as-a-Judge Evaluation Mechanism.}
Due to the limitations of traditional string-matching metrics in evaluating complex multimodal reasoning, we implement a robust LLM-as-a-Judge protocol using DeepSeek-V3.2. The judge is provided with the Objective GT and evaluates the model outputs across two orthogonal dimensions:
\begin{itemize}
    \item \textbf{Fidelity (F) [0-5]:} Measures the strict adherence to visual facts. Penalizes the model for fabricating entities, imposing fake causality, or suffering from mechanistic mode collapse.
    \item \textbf{Resilience (R) [0-5]:} Measures the completeness of information extraction. Penalizes the model for being hijacked by adversarial text, omitting crucial visual details, or triggering unjustified epistemic refusal.
\end{itemize}

\begin{table*}[t]
\centering
\small 
\caption{Comprehensive Evaluation on the ModeVent Benchmark. We report Fidelity (F), Resilience (R), and Total Scores across 7 major hallucination categories. The results are further stratified by semantic distance: \textbf{Inliers} (In-Domain interference) and \textbf{Outliers} (Out-of-Domain irrelevance). The best results in each comparison are highlighted in \textbf{bold}.}
\label{tab:main_results}
\begin{tabular}{ll ccc ccc ccc}
\toprule
\multirow{2}{*}{\textbf{Error Category}} & \multirow{2}{*}{\textbf{Data Label}} & \multicolumn{3}{c}{\textbf{Baseline}} & \multicolumn{3}{c}{\textbf{MODE-RAG (Ours)}} & \multicolumn{3}{c}{\textbf{Improvement ($\Delta$)}} \\
\cmidrule(lr){3-5} \cmidrule(lr){6-8} \cmidrule(lr){9-11}
& & F & R & Tot & F & R & Tot & $\Delta$F & $\Delta$R & $\Delta$Tot \\
\midrule
\multirow{3}{*}{Attribute Hijacking}
& Inliers                 & 1.45 & 0.85 & 2.30 & \textbf{2.40} & \textbf{1.26} & \textbf{3.66} & +0.95 & +0.41 & +1.36 \\
& Outliers                & 1.91 & 1.34 & 3.25 & \textbf{2.38} & \textbf{1.64} & \textbf{4.02} & +0.47 & +0.30 & +0.77 \\
& \textit{Overall}        & 1.66 & 1.08 & 2.74 & \textbf{2.39} & \textbf{1.44} & \textbf{3.82} & +0.72 & +0.36 & +1.08 \\
\addlinespace
\multirow{3}{*}{Causal Imposition}
& Inliers                 & 1.82 & \textbf{1.78} & 3.60 & \textbf{2.64} & 1.39 & \textbf{4.03} & +0.82 & -0.39 & +0.43 \\
& Outliers                & 1.86 & 1.93 & 3.79 & \textbf{2.78} & \textbf{2.19} & \textbf{4.97} & +0.92 & +0.26 & +1.18 \\
& \textit{Overall}        & 1.84 & \textbf{1.86} & 3.70 & \textbf{2.71} & 1.81 & \textbf{4.52} & +0.87 & -0.05 & +0.82 \\
\addlinespace
\multirow{3}{*}{Information Sparsity}
& Inliers                 & 2.32 & 1.61 & 3.92 & \textbf{3.14} & \textbf{1.71} & \textbf{4.86} & +0.83 & +0.11 & +0.93 \\
& Outliers                & 2.05 & 1.88 & 3.93 & \textbf{3.60} & \textbf{2.10} & \textbf{5.71} & +1.55 & +0.22 & +1.78 \\
& \textit{Overall}        & 2.20 & 1.72 & 3.93 & \textbf{3.34} & \textbf{1.88} & \textbf{5.22} & +1.14 & +0.16 & +1.30 \\
\addlinespace
\multirow{3}{*}{Majority Text Bias}
& Inliers                 & 2.83 & \textbf{2.98} & 5.82 & \textbf{3.40} & 2.83 & \textbf{6.23} & +0.57 & -0.15 & +0.42 \\
& Outliers                & 2.43 & 2.61 & 5.04 & \textbf{3.91} & \textbf{3.45} & \textbf{7.36} & +1.48 & +0.84 & +2.31 \\
& \textit{Overall}        & 2.62 & 2.79 & 5.41 & \textbf{3.67} & \textbf{3.16} & \textbf{6.83} & +1.05 & +0.37 & +1.42 \\
\addlinespace
\multirow{3}{*}{Metadata Redundancy}
& Inliers                 & 2.94 & \textbf{2.32} & 5.26 & \textbf{3.80} & 2.02 & \textbf{5.82} & +0.86 & -0.30 & +0.56 \\
& Outliers                & 2.53 & 2.41 & 4.94 & \textbf{3.71} & \textbf{2.77} & \textbf{6.49} & +1.19 & +0.36 & +1.54 \\
& \textit{Overall}        & 2.73 & 2.37 & 5.10 & \textbf{3.76} & \textbf{2.40} & \textbf{6.16} & +1.03 & +0.04 & +1.07 \\
\addlinespace
\multirow{3}{*}{Out-of-Domain Irrelevance}
& Inliers                 & 2.86 & \textbf{2.19} & 5.05 & \textbf{3.31} & 1.94 & \textbf{5.25} & +0.45 & -0.25 & +0.20 \\
& Outliers                & 2.75 & 2.72 & 5.46 & \textbf{4.00} & \textbf{3.14} & \textbf{7.14} & +1.25 & +0.42 & +1.68 \\
& \textit{Overall}        & 2.80 & 2.47 & 5.27 & \textbf{3.67} & \textbf{2.57} & \textbf{6.24} & +0.87 & +0.10 & +0.98 \\
\addlinespace
\multirow{3}{*}{Scene Misalignment}
& Inliers                 & 2.56 & \textbf{2.13} & 4.69 & \textbf{2.89} & 1.90 & \textbf{4.79} & +0.32 & -0.23 & +0.10 \\
& Outliers                & 2.61 & 2.29 & 4.90 & \textbf{3.30} & \textbf{2.74} & \textbf{6.04} & +0.70 & +0.45 & +1.14 \\
& \textit{Overall}        & 2.59 & 2.21 & 4.80 & \textbf{3.11} & \textbf{2.34} & \textbf{5.45} & +0.52 & +0.13 & +0.65 \\
\midrule
\multirow{3}{*}{\textbf{Average}}
& \textbf{Inliers}        & 2.37 & \textbf{1.94} & 4.31 & \textbf{3.07} & 1.84 & \textbf{4.91} & +0.70 & -0.10 & +0.60 \\
& \textbf{Outliers}       & 2.31 & 2.18 & 4.50 & \textbf{3.39} & \textbf{2.59} & \textbf{5.98} & +1.07 & +0.41 & +1.48 \\
& \textbf{\textit{Overall}} & 2.34 & 2.06 & 4.40 & \textbf{3.23} & \textbf{2.22} & \textbf{5.45} & +0.89 & +0.16 & +1.04 \\
\bottomrule
\end{tabular}
\end{table*}

\subsection{Main Results and Quantitative Analysis}
As shown in Table \ref{tab:main_results}, MODE-RAG significantly and consistently outperforms the Baseline across all 7 RAG error categories, achieving a global \textbf{Average Total Score improvement of +1.04} (from 4.40 to 5.45). The dual-dimension analysis reveals that our system successfully resolves the intervention paradox by boosting Fidelity ($\Delta$F = +0.89) without sacrificing information extraction ($\Delta$R = +0.16).

\paragraph{Conquering Outliers Hijacking.}
In Outliers scenarios, traditional RAG models suffer from severe "Attention Hijacking," where the LLM abandons visual evidence to blindly follow irrelevant or malicious text. Our results show that MODE-RAG excels in these extreme conditions, yielding a massive $\Delta$Total improvement of \textbf{+1.48}. The most striking gains are observed in \textit{Majority Text Bias} ($\Delta$Total = +2.31) and \textit{Out-of-Domain Irrelevance} ($\Delta$Total = +1.68). This validates the efficacy of our \textbf{Ret-Agent}. By explicitly calculating the manifold distance between the text and the \textit{Visual Logic Graph}, the system accurately detects epistemic uncertainty and triggers the \texttt{[EMPTY CONTEXT FALLBACK]}, forcing the model to anchor its generation purely on the physical visual evidence rather than fabricated text.

\paragraph{Refining Inliers Extraction.}
Inliers scenarios present a highly nuanced challenge: the retrieved text is semantically relevant but contains redundant metadata or slightly conflicting attributes. A naive filtering approach often leads to unjustified refusal, resulting in low Resilience. However, MODE-RAG achieves a +0.60 $\Delta$Total improvement in Inliers cases. Notably, in the \textit{Information Sparsity} category, our model achieves a significant $\Delta$Total of +0.93. This demonstrates the success of the \textbf{Smart Synthesis} protocol within the Gen-Agent, which safely fuses domain-specific nouns from the text (e.g., specific names or medical terms) with the MCTS-verified visual actions, thereby preserving rich background context without hallucinating actions.

\begin{figure*}[t]
\centering
\includegraphics[width=\textwidth]{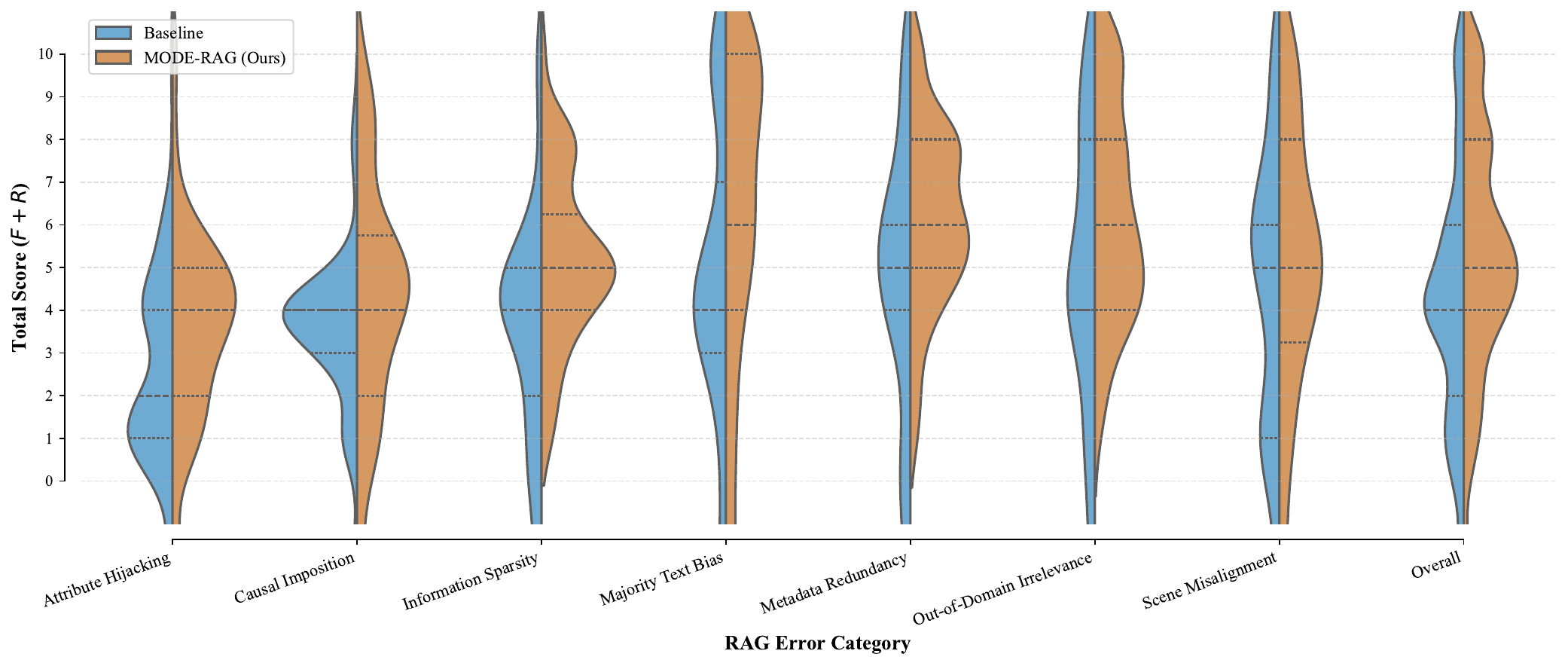}
\caption{Split violin plots of Total Scores (Fidelity + Resilience) across seven RAG error categories and overall performance. The left (blue) and right (orange) distributions represent the Baseline and MODE-RAG, respectively. Dotted lines indicate the median and interquartile ranges. MODE-RAG significantly suppresses zero-score catastrophic failures and shifts the performance mass towards high-fidelity regions.}
\label{fig:score_dist}
\end{figure*}

\paragraph{Performance Stability and Failure Suppression.} 
While Table \ref{tab:main_results} demonstrates mean improvements, Figure \ref{fig:score_dist} provides a deeper look into the system's robustness by visualizing the score distribution. 
A critical observation is the suppression of ``catastrophic failures'' in the 0–2 score range. 
In categories like \textit{Majority Text Bias} and \textit{Metadata Redundancy}, the Baseline distribution exhibits a significant density bulge at the bottom, corresponding to cases where the model suffered from severe mode collapse (e.g., stuttering loops) or total attention hijacking. 
In contrast, MODE-RAG's distribution is markedly narrower at the base, effectively establishing a ``safety floor'' through the \textbf{Dead Man's Switch} and \textbf{MCTS pruning} mechanisms.

Furthermore, the \textit{Overall} density for MODE-RAG shows a decisive upward shift, with the median score and interquartile ranges positioned substantially higher than the Baseline. 
This shift is most prominent in \textit{Out-of-Domain Irrelevance}, where MODE-RAG transforms a low-fidelity bimodal distribution into a concentrated high-score peak. 
This proves that the \textbf{FE-Router} correctly identifies high-uncertainty scenarios, allowing the multi-agent pipeline to neutralize adversarial noise and anchor the final generation to the physically-grounded visual logic.

While the results above confirm that MODE-RAG consistently outperforms the vanilla foundational kernel, we further evaluate our framework against alternative mitigation paradigms to ensure a thorough assessment. The full benchmarking results across all five methods (Vanilla Baseline, Self-RAG, SelfCheckGPT, Woodpecker, and MODE-RAG) on the ModeVent dataset are detailed in Appendix \ref{sec:appendix_additional_backbones}.

\subsection{Ablation on Mechanistic Failures}
Beyond semantic conflicts, our error logs revealed that lightweight LLM kernels frequently suffer from mechanistic failures under adversarial stress. We observed two primary collapse patterns in the Baseline: \textit{Mode Collapse} (e.g., severe stuttering loops like "even even even") and \textit{Prompt Bleed-through} (leaking internal system tags or metadata like "addCriterion"). These failures historically resulted in 0-point scores for Fidelity. By incorporating an internal, rule-based \textbf{Dead Man's Switch} within the Gen-Agent---a deterministic regular-expression interceptor, MODE-RAG effectively establishes a safety floor. This mechanism successfully neutralizes catastrophic formatting failures, seamlessly downgrading to a safe textual reading-comprehension state when the VLM's predictive coding collapses. 

To explicitly demonstrate how our decoupled architecture resolves the intervention paradox in practice, we provide a detailed comparative case study of four adversarial testing scenarios in \textbf{Appendix \ref{sec:appendix_case_study}}.

\subsection{Computational Efficiency Analysis}
\label{sec:efficiency}

To evaluate the practical deployability of MODE-RAG, we analyze its computational overhead against the Vanilla M-RAG baseline across the 1,000 video queries in the ModeVent benchmark. On average, the baseline foundational kernel requires 18.5 seconds to process a single multimodal query. In comparison, due to the multi-agent orchestration and MCTS-guided test-time reasoning scaling, MODE-RAG increases the average processing time to 26.2 seconds per query. This represents a moderate 1.42$\times$ increase in time consumption, translating to approximately 7.3 hours of execution time when evaluating the entire benchmark sequentially on a single-threaded pipeline. It is worth noting that because the stage-specific agent interventions and evaluation queries are inherently decoupled, this computational overhead can be significantly mitigated through standard multi-threading, asynchronous scheduling, and parallel execution techniques in production environments.

\section{Conclusions}
In this paper, we proposed MODE-RAG, a mechanistically grounded multi-agent framework that addresses the intervention paradox in multimodal RAG systems by dynamically gating interventions through a router driven by Variational Free Energy (VFE) and internal attention states (ATLAS). By categorizing hallucinations into nine distinct types across the system's lifecycle, we developed specialized agents---integrating Monte Carlo Tree Search (MCTS) for causal derivation and logit perturbations for sycophancy suppression---to ensure factual grounding and logical consistency. Furthermore, we introduced ModeVent, a targeted benchmark designed to evaluate system susceptibility to manifold outliers and complex visual-textual conflicts. Experimental results demonstrate that MODE-RAG effectively reduces hallucination rates and enhances the structural stability of M-RAG systems, providing a robust and scalable solution for reliable multimodal reasoning.

\section*{Acknowledgment}
This work was supported by the Ministry of Science and Technology of China under Grant No. 2025ZD0123800, the HUST Interdisciplinary Research Program under Grant No. 2025JCYJ077, and the KingSoft 2026 University-Industry Project. 

\bibliography{custom}

\appendix
\section*{Appendix} 

\renewcommand{\thetable}{A\arabic{table}}
\setcounter{table}{0}

\renewcommand{\thefigure}{A\arabic{figure}}
\setcounter{figure}{0}

\section{Case Studies}
\label{sec:appendix}

\subsection*{Mechanistic Analysis of System Interventions}
\label{sec:appendix_case_study}

\begin{table*}[t]
\setlength{\cmidrulewidth}{0.15pt}
\centering
\small
\renewcommand{\arraystretch}{1.1}
\newcommand{\tightdoubleline}{\specialrule{0.4pt}{1.5pt}{0.3pt}\specialrule{0.4pt}{0.3pt}{1.5pt}}

\caption{Comparative analysis of system outputs across four adversarial testing scenarios. Key hallucinations and mode collapses in the Baseline are highlighted in \textbf{bold}. Case separation is enforced using heavy rules to maintain visual hierarchy.}
\label{tab:case_study}
\begin{tabularx}{\textwidth}{@{} >{\raggedright\arraybackslash}X >{\raggedright\arraybackslash}X @{}}
\toprule
\addlinespace[1ex]
\multicolumn{2}{@{}l@{}}{\textbf{Scenario 1: Physical Scene Replacement}} \\ 
\multicolumn{2}{@{}p{\textwidth}@{}}{\textbf{Query:} What is the person in the video doing? What is the environment like?} \\
\multicolumn{2}{@{}p{\textwidth}@{}}{\textbf{Retrieved Text:} \textit{[4 adversarial chunks describing intense water sports: surfing, floods, etc.]}} \\
\addlinespace[1ex]
\textbf{Baseline (Vanilla M-RAG)} & \textbf{MODE-RAG (Ours)} \\
\cmidrule(r){1-1} \cmidrule(l){2-2}
The person in the video appears to be engaging in \textbf{snowshoeing or cross-country skiing} in a snowy forested area... & The person in the video is \textbf{standing on a snowboard, preparing to descend a snowy slope}. The environment appears to be a snowy landscape... \\

\addlinespace[1ex]
\tightdoubleline
\addlinespace[1ex]

\multicolumn{2}{@{}l@{}}{\textbf{Scenario 2: Micro-Detail Tampering}} \\
\multicolumn{2}{@{}p{\textwidth}@{}}{\textbf{Query:} What equipment is the person using in the snow?} \\
\multicolumn{2}{@{}p{\textwidth}@{}}{\textbf{Retrieved Text:} \textit{[3 chunks describing "twin-tip skis," "bright red skis," and "carbon fiber poles."]}} \\
\addlinespace[1ex]
\textbf{Baseline (Vanilla M-RAG)} & \textbf{MODE-RAG (Ours)} \\
\cmidrule(r){1-1} \cmidrule(l){2-2}
The person in the video is using \textbf{twin-tip skis and carbon fiber poles}. & The person is using \textbf{snowshoes} in the snow. \\

\addlinespace[1ex]
\tightdoubleline
\addlinespace[1ex]

\multicolumn{2}{@{}l@{}}{\textbf{Scenario 3: High-Order Causal Imposition}} \\
\multicolumn{2}{@{}p{\textwidth}@{}}{\textbf{Query:} What equipment is the person using in the snow?} \\
\multicolumn{2}{@{}p{\textwidth}@{}}{\textbf{Retrieved Text:} \textit{[Chunks describing competitive winter sports, Olympic trials, and avalanche events.]}} \\
\addlinespace[1ex]
\textbf{Baseline (Vanilla M-RAG)} & \textbf{MODE-RAG (Ours)} \\
\cmidrule(r){1-1} \cmidrule(l){2-2}
The person is \textbf{performing a backflip}... practicing for the Winter Olympics, \textbf{attempting a triple cork 1440 spin}... actually \textbf{outrunning an avalanche}. & The specific action appears to be preparing for winter sports... \textbf{Insufficient information to determine the background of this event}. \\

\addlinespace[1ex]
\tightdoubleline
\addlinespace[1ex]

\multicolumn{2}{@{}l@{}}{\textbf{Scenario 4: Majority Text Bias}} \\
\multicolumn{2}{@{}p{\textwidth}@{}}{\textbf{Query:} What is the person doing and where is this filmed?} \\
\multicolumn{2}{@{}p{\textwidth}@{}}{\textbf{Retrieved Text:} \textit{[Mixed noise: 3 water-sports chunks vs. 1 relevant "Swiss Alps" snowboarding chunk.]}} \\
\addlinespace[1ex]
\textbf{Baseline (Vanilla M-RAG)} & \textbf{MODE-RAG (Ours)} \\
\cmidrule(r){1-1} \cmidrule(l){2-2}
The person in the video is backcountry snowboarding... The location is the Swiss Alps, which is known for its its \textbf{[Followed by stuttering mode collapse]}. & \textbf{The person is preparing to descend a snowy slope on a snowboard.} This activity is filmed in the \textbf{Swiss Alps, specifically in a pristine snowy forest}. \\

\addlinespace[1ex]
\bottomrule
\end{tabularx}
\end{table*}

This appendix provides additional qualitative evidence on how the decoupled architecture resolves the intervention paradox, we present a comparative analysis of four adversarial queries from the ModeVent benchmark test logs. 

\paragraph{Objective Visual Ground Truth.} Across all four test scenarios detailed in Table \ref{tab:case_study}, the underlying visual evidence remains constant: the video depicts a person in a serene, snowy forest, either wearing snowshoes or standing still on a snowboard preparing to descend. There are no extreme stunts, competitive sporting events, or water-related elements present in the actual footage.

\setcounter{table}{0}
\begin{table*}[!t]
\renewcommand{\thetable}{B\arabic{table}}
\centering
\small 
\caption{Comprehensive Evaluation on the ModeVent Benchmark. We report Fidelity (F), Resilience (R), and Total Scores across 7 major hallucination categories, further stratified by semantic distance: \textbf{Inliers} and \textbf{Outliers}. Notably, we introduce the Video-adapted \textbf{Woodpecker}, the text-based \textbf{SelfCheckGPT}, and \textbf{Self-RAG} as competitive baselines. Despite their strong performance, our proposed \textbf{MODE-RAG} maintains a clear advantage across the majority of metrics and scenarios. The best results in each comparison are highlighted in \textbf{bold}.}
\label{tab:main_results_horizontal}
\begin{tabular}{ll ccc ccc ccc}
\toprule
\multirow{2}{*}{\textbf{Error Category}} & \multirow{2}{*}{\textbf{Method}} & \multicolumn{3}{c}{\textbf{Inliers}} & \multicolumn{3}{c}{\textbf{Outliers}} & \multicolumn{3}{c}{\textbf{\textit{Overall}}} \\
\cmidrule(lr){3-5} \cmidrule(lr){6-8} \cmidrule(lr){9-11}
& & F & R & Total & F & R & Total & F & R & Total \\
\midrule
\multirow{5}{*}{Attribute Hijacking}
& Baseline         & 1.45 & 0.85 & 2.30 & 1.91 & 1.34 & 3.25 & 1.66 & 1.08 & 2.74 \\
& SelfCheckGPT     & 1.10 & 0.22 & 1.32 & 1.29 & 0.35 & 1.64 & 1.19 & 0.28 & 1.47 \\
& Self-RAG         & 2.23 & 1.29 & 3.52 & \textbf{2.70} & 1.89 & 4.59 & \textbf{2.44} & 1.56 & \textbf{4.01} \\
& Woodpecker       & 1.90 & \textbf{1.36} & 3.26 & 2.56 & \textbf{2.15} & \textbf{4.71} & 2.20 & \textbf{1.72} & 3.93 \\
& \textbf{Ours}    & \textbf{2.40} & 1.26 & \textbf{3.66} & 2.38 & 1.64 & 4.02 & 2.39 & 1.44 & 3.82 \\
\addlinespace
\multirow{5}{*}{Causal Imposition}
& Baseline         & 1.82 & \textbf{1.78} & 3.60 & 1.86 & 1.93 & 3.79 & 1.84 & 1.86 & 3.70 \\
& SelfCheckGPT     & 1.07 & 0.45 & 1.52 & 1.00 & 0.40 & 1.40 & 1.03 & 0.42 & 1.46 \\
& Self-RAG         & \textbf{2.76} & 1.63 & \textbf{4.39} & 3.05 & 2.17 & 5.23 & \textbf{2.91} & 1.91 & 4.82 \\
& Woodpecker       & 2.25 & 1.76 & 4.01 & \textbf{3.30} & \textbf{3.07} & \textbf{6.37} & 2.80 & \textbf{2.44} & \textbf{5.24} \\
& \textbf{Ours}    & 2.64 & 1.39 & 4.03 & 2.78 & 2.19 & 4.97 & 2.71 & 1.81 & 4.52 \\
\addlinespace
\multirow{5}{*}{Information Sparsity}
& Baseline         & 2.32 & 1.61 & 3.92 & 2.05 & 1.88 & 3.93 & 2.20 & 1.72 & 3.93 \\
& SelfCheckGPT     & \textbf{4.80} & 1.16 & \textbf{5.96} & \textbf{4.35} & 1.44 & \textbf{5.79} & \textbf{4.60} & 1.28 & \textbf{5.89} \\
& Self-RAG         & 2.95 & 1.10 & 4.05 & 2.63 & 1.48 & 4.11 & 2.81 & 1.27 & 4.08 \\
& Woodpecker       & 2.20 & 1.30 & 3.51 & 2.75 & \textbf{2.15} & 4.90 & 2.44 & 1.67 & 4.12 \\
& \textbf{Ours}    & 3.14 & \textbf{1.71} & 4.86 & 3.60 & 2.10 & 5.71 & 3.34 & \textbf{1.88} & 5.22 \\
\addlinespace
\multirow{5}{*}{Majority Text Bias}
& Baseline         & 2.83 & \textbf{2.98} & 5.82 & 2.43 & 2.61 & 5.04 & 2.62 & 2.79 & 5.41 \\
& SelfCheckGPT     & 1.61 & 1.20 & 2.80 & 1.25 & 1.01 & 2.26 & 1.41 & 1.09 & 2.50 \\
& Self-RAG         & 3.21 & 2.38 & 5.59 & 3.78 & 3.18 & 6.96 & 3.53 & 2.83 & 6.35 \\
& Woodpecker       & 3.02 & 2.35 & 5.37 & 3.11 & 2.76 & 5.87 & 3.07 & 2.58 & 5.65 \\
& \textbf{Ours}    & \textbf{3.40} & 2.83 & \textbf{6.23} & \textbf{3.91} & \textbf{3.45} & \textbf{7.36} & \textbf{3.67} & \textbf{3.16} & \textbf{6.83} \\
\addlinespace
\multirow{5}{*}{Metadata Redundancy}
& Baseline         & 2.94 & \textbf{2.32} & 5.26 & 2.53 & 2.41 & 4.94 & 2.73 & 2.37 & 5.10 \\
& SelfCheckGPT     & \textbf{3.99} & 2.31 & \textbf{6.29} & 3.23 & 2.21 & 5.44 & 3.61 & 2.26 & 5.86 \\
& Self-RAG         & 3.01 & 2.03 & 5.04 & 3.51 & 2.74 & 6.25 & 3.26 & 2.39 & 5.65 \\
& Woodpecker       & 2.35 & 1.71 & 4.06 & 2.75 & 2.63 & 5.38 & 2.55 & 2.18 & 4.73 \\
& \textbf{Ours}    & 3.80 & 2.02 & 5.82 & \textbf{3.71} & \textbf{2.77} & \textbf{6.49} & \textbf{3.76} & \textbf{2.40} & \textbf{6.16} \\
\addlinespace
\multirow{5}{*}{Out-of-Domain Irrelevance}
& Baseline         & 2.86 & 2.19 & 5.05 & 2.75 & 2.72 & 5.46 & 2.80 & 2.47 & 5.27 \\
& SelfCheckGPT     & 1.34 & 0.23 & 1.56 & 2.14 & 0.41 & 2.55 & 1.75 & 0.32 & 2.06 \\
& Self-RAG         & \textbf{3.55} & \textbf{2.21} & \textbf{5.76} & 3.55 & 2.74 & 6.29 & 3.55 & 2.48 & 6.03 \\
& Woodpecker       & 3.01 & 2.06 & 5.07 & 3.24 & 2.97 & 6.21 & 3.13 & 2.52 & 5.64 \\
& \textbf{Ours}    & 3.31 & 1.94 & 5.25 & \textbf{4.00} & \textbf{3.14} & \textbf{7.14} & \textbf{3.67} & \textbf{2.57} & \textbf{6.24} \\
\addlinespace
\multirow{5}{*}{Scene Misalignment}
& Baseline         & 2.56 & 2.13 & 4.69 & 2.61 & 2.29 & 4.90 & 2.59 & 2.21 & 4.80 \\
& SelfCheckGPT     & 1.05 & 0.02 & 1.06 & 1.03 & 0.07 & 1.10 & 1.04 & 0.05 & 1.08 \\
& Self-RAG         & \textbf{3.29} & \textbf{2.26} & \textbf{5.55} & 3.27 & 2.42 & 5.70 & \textbf{3.28} & 2.34 & \textbf{5.63} \\
& Woodpecker       & 2.62 & 1.94 & 4.56 & 3.03 & \textbf{2.89} & 5.91 & 2.84 & \textbf{2.44} & 5.27 \\
& \textbf{Ours}    & 2.89 & 1.90 & 4.79 & \textbf{3.30} & 2.74 & \textbf{6.04} & 3.11 & 2.34 & 5.45 \\
\midrule
\multirow{5}{*}{\textbf{Average}}
& Baseline     & 2.37 & \textbf{1.94} & 4.31 & 2.31 & 2.18 & 4.50 & 2.34 & 2.06 & 4.40 \\
& SelfCheckGPT & 2.20 & 0.80 & 3.00 & 1.99 & 0.84 & 2.83 & 2.10 & 0.82 & 2.92 \\
& Self-RAG     & 2.98 & 1.80 & 4.78 & 3.24 & 2.41 & 5.64 & 3.11 & 2.11 & 5.21 \\
& Woodpecker   & 2.45 & 1.75 & 4.21 & 2.97 & \textbf{2.68} & 5.65 & 2.71 & \textbf{2.22} & 4.93 \\
& \textbf{Ours}& \textbf{3.07} & 1.84 & \textbf{4.91} & \textbf{3.39} & 2.59 & \textbf{5.98} & \textbf{3.23} & \textbf{2.22} & \textbf{5.45} \\
\bottomrule
\end{tabular}
\end{table*}

\paragraph{Combating Attribute Hijacking and Perception Omission.} 
Scenarios 1 and 2 highlight the Baseline's vulnerability to semantic coercion. Despite the visual evidence clearly showing a snowboard or snowshoes, the injection of text describing ``cross-country skiing'' or ``twin-tip skis'' hijacked the Baseline's attention, causing it to blindly hallucinate equipment not present in the video. In contrast, MODE-RAG's \textbf{Per-Agent} enforces a strict ``visual-first'' extraction. By isolating atomic visual facts before textual integration, the system successfully overrides the adversarial text, accurately maintaining the physical reality of the scene.

\paragraph{Suppressing Sycophancy and Causal Fabrication.}
Scenario 3 demonstrates a severe case of Causal Imposition. Confronted with text describing competitive winter sports, the Baseline model exhibits extreme sycophancy, inventing a massive, Hollywood-style narrative involving a ``triple cork 1440 spin'' and ``outrunning an avalanche.'' This exposes the danger of unguided LLM reasoning, where the model prioritizes narrative alignment with the text over physical constraints. MODE-RAG neutralizes this through the \textbf{Rea-Agent's MCTS DAG}. Since an avalanche or a backflip cannot be topologically derived from the Per-Agent's root node (standing still), the MCTS prunes these branches entirely, allowing the Gen-Agent to safely output a justified epistemic refusal regarding the background context.

\paragraph{Preventing Mechanistic Mode Collapse.}
Scenario 4 exposes a critical physical limitation of lightweight LLM kernels. When subjected to Majority Text Bias (a 3:1 ratio of water-sports noise to relevant snow text), the Baseline model's attention mechanism collapses under the conflicting semantic density, resulting in a stuttering loop and system paralysis. MODE-RAG bypasses this failure mode completely. Prior to generation, the \textbf{Ret-Agent} actively computes the manifold distance between the visual log and the candidate texts, discarding the three contradictory water-sports chunks upfront. This listwise cross-check purifies the context window, feeding the generator a clean, aligned prompt that guarantees formatting stability and flawless factual synthesis.

\section{Results on Additional Backbones}
\label{sec:appendix_additional_backbones}

To comprehensively verify the effectiveness of the MODE-RAG framework, we conduct an extensive comparative analysis against multiple established mitigation paradigms in recent literature. Specifically, our benchmark encompasses a total of five distinct methodological configurations:
\begin{enumerate}
    \item \textbf{Vanilla M-RAG (Baseline):} The foundational unguided VLM (Qwen-2.5-VL-7B) executing direct multimodal generation.
    \item \textbf{Self-RAG}~\cite{asai2024self}: An end-to-end framework that trains the model to self-reflect on retrieved passages and generations via reflection tokens.
    \item \textbf{SelfCheckGPT}~\cite{manakul-etal-2023-selfcheckgpt}: A zero-resource sampling-based approach that detects hallucinations via stochastic consistency checks.
    \item \textbf{Woodpecker}~\cite{yin2024woodpecker}: A training-free, post-hoc correction pipeline designed to rectify multi-modal fabrications through diagnostic querying.
    \item \textbf{MODE-RAG (Ours):} Our proposed hierarchical, variational free energy-gated multi-agent intervention framework.
\end{enumerate}

Table \ref{tab:main_results_horizontal} presents the full quantitative comparison across these five methods on the polar extremes of the ModeVent dataset.

\subsection{Implementation of Additional Baselines}

While Vanilla M-RAG requires no architectural modification and MODE-RAG is detailed in Section 4, the remaining three baselines (Woodpecker, SelfCheckGPT, and Self-RAG) were originally developed for static images or pure text. Below, we outline the specific multimodal adaptations and pipeline configurations required to deploy them within our adversarial video RAG setting.

\paragraph{Video-Adapted Woodpecker.}
We adapt the Woodpecker framework~\cite{yin2024woodpecker}—initially an image-centric hallucination corrector—to the video domain by shifting the focus from spatial object misidentification to temporal dynamics (e.g., fabricated actions or incorrect event sequences). The adapted pipeline operates in four stages: (1) Drafting: A standard multimodal RAG setup generates an initial answer. (2) Question Generation: An LLM extracts action-centric claims and temporal events from the draft, formulating targeted verification questions. (3) Visual Verification: A Video-LLM acts as an independent visual expert. Crucially, external texts are masked to ensure the model relies solely on raw video frames for objective fidelity. (4) Correction: The verified answers form a Visual Fact-Sheet, guiding the LLM to revise the initial draft and prune spatiotemporal hallucinations.

\paragraph{Multimodal SelfCheckGPT.}
To complement the visual-centric verification, we implement an alternative, uncertainty-based baseline by adapting SelfCheckGPT~\cite{manakul-etal-2023-selfcheckgpt} from black-box text evaluation to the multimodal RAG domain. This zero-shot pipeline addresses adversarial textual noise through generation consistency, executing in three stages: (1) Multi-Sample Generation: Multiple independent candidate answers are generated using high-temperature sampling. (2) Consistency Voting: Instead of standard token-level probability checks, a semantic overlap metric identifies the most frequent consensus among the candidates. (3) Refinement: The LLM acts as a strict validator, cross-referencing the candidate consensus against the raw retrieved texts to synthesize a final factual response. Additionally, we integrate a dynamic memory-recovery mechanism with progressive token-throttling to handle potential Out-Of-Memory errors during large-scale evaluation.

\paragraph{Multimodal Self-RAG.}
Given that the original Self-RAG~\cite{asai2024self} is a text-to-text framework designed to critique retrieved textual passages, we adapt it for video reasoning via a two-stage cascaded pipeline. This approach bridges the modality gap while preserving the model's reflective capabilities: (1) Visual Translation: A Vision-Language Model first processes the raw video frames alongside the adversarial retrieved contexts to generate a comprehensive text-based description of the visual scenes, actions, and objects. (2) Reflective Generation: This visual description is subsequently injected into the Self-RAG model using its native retrieval syntax. Treating this textual translation as the primary retrieved evidence, the Self-RAG model leverages its intrinsic reflection tokens to evaluate the fidelity of the provided information and synthesize the final answer to the user's query.

\subsection{Result Analysis and Discussion}
\label{sec:appendix_discussion}

The comprehensive empirical results presented in Table \ref{tab:main_results_horizontal} demonstrate the performance trade-offs, highlighting both the global strengths and the localized limitations of the proposed MODE-RAG framework.

\paragraph{Overall Strengths and Outlier Robustness.}
MODE-RAG achieves the highest global performance with an \textit{Overall} Average Total Score of \textbf{5.45}, consistently outperforming all four competitive baselines (Baseline: 4.40, SelfCheckGPT: 2.92, Self-RAG: 5.21, Woodpecker: 4.93). The primary architectural advantage of our framework lies in its exceptional robustness against \textbf{Outliers (Hard-OOD)} scenarios, where it reaches an average total score of \textbf{5.98}. Specifically, in categories heavily plagued by aggressive external text noise—such as \textit{Majority Text Bias} (7.36) and \textit{Out-of-Domain Irrelevance} (7.14)—MODE-RAG delivers a substantial performance leap. This consistently validates the efficacy of our thermodynamic gating via the FE-Router and the manifold filtering via the Ret-Agent. By proactively evaluating the epistemic uncertainty and discarding highly mismatched text chunks upfront, our system effectively prevents the LLM kernel from experiencing attention hijacking, thereby securing a strong safety floor for factual cross-modal synthesis.

\paragraph{Vulnerability to Information Sparsity.}
Despite its global superiority, the multi-agent execution within MODE-RAG exhibits localized deficits under specific error contexts. In the \textit{Information Sparsity} category, MODE-RAG (Overall Total: 5.22) is noticeably outperformed by the text-based SelfCheckGPT, which achieves a dominant score of \textbf{5.89}. This deficit occurs because when the retrieved context is extremely sparse, SelfCheckGPT’s high-temperature multi-sample consistency voting natively excels at consensus-driven extraction. In contrast, our rigid multi-agent validation schema can occasionally become overly restrictive, leading to redundant processing steps without gaining an additional informative edge.

\paragraph{Conservative Pruning in Complex Reasoning.}
Another limitation is observed in the \textit{Causal Imposition} category, where Woodpecker outperforms our method in both Outliers (6.37 vs. 4.97) and Overall (5.24 vs. 4.52) metrics. A granular examination reveals that this is primarily driven by a drop in our Resilience (R) scores (1.81 vs. Woodpecker's 2.44). Because Woodpecker leverages an aggressive post-hoc prompt rewriting strategy based on direct question-answering, it forces the model to actively correct claims. MODE-RAG, conversely, relies on a strict MCTS causal DAG; when a claim cannot be topologically derived from the visual invariants, the system tends to trigger a conservative \textit{Epistemic Refusal} (i.e., acknowledging insufficient information). While this strictness preserves visual Fidelity, it inherently sacrifices descriptive completeness (Resilience) when facing high-order causal fabrications.

\section{Data Construction Examples}
To automate the construction of the ModeVent benchmark, we leveraged \textbf{DeepSeek-V3.2} to synthesize adversarial test scenarios from MultiVent’s ground truth. These misleading queries are strategically designed to reflect the hallucination taxonomy introduced in \cref{sec:intro}, ensuring a comprehensive evaluation of model vulnerabilities. In this section, we present representative examples of the challenging queries generated through this pipeline.

\vspace{1em}

\begin{tcolorbox}[
    colback=gray!5!white,      
    colframe=gray!60!black,    
    title={Example 1: Causal Imposition}, 
    fonttitle=\bfseries,       
    breakable,                 
    boxrule=0.8pt,            
    arc=3pt,                  
    left=4pt, right=4pt, top=4pt, bottom=4pt 
]

\textbf{Ground Truth:} \\
This is a news report from TVBS News about a medical condition called \"cytokine storm,\" which can be fatal. The report features interviews with doctors from Taipei Veterans General Hospital and a nutritionist, who discuss how this immune overreaction can damage organs like the lungs, as shown by X-ray images. Experts advise that during the pandemic, people should manage stress, get enough sleep, and maintain a balanced diet to strengthen their immune systems. The segment also shows scenes of public health measures, including disinfection at a train station and people receiving vaccinations.''

\tcblower 

\textbf{Error Type:} Causal Imposition \\[0.5em]
\textbf{Generated Query:} \\
What is being discussed in the news report, and what is the purported origin of the medical condition according to the background context?

\end{tcolorbox}

\vspace{0.5em}

\begin{tcolorbox}[
    colback=gray!5!white,      
    colframe=gray!60!black,    
    title={Example 2: Attribute Hijacking}, 
    fonttitle=\bfseries,       
    breakable,                 
    boxrule=0.8pt,            
    arc=3pt,                  
    left=4pt, right=4pt, top=4pt, bottom=4pt 
]

\textbf{Ground Truth:} \\
The video features an interview with Professor Chen Jian from Beijing Hospital's Hepatobiliary Surgery Department. He discusses the treatment of liver abscesses, explaining that early-stage cases are often managed conservatively with antibiotics, while larger abscesses may require drainage or surgical removal. The conversation takes place in an office setting with bookshelves in the background, and animated graphics illustrate medical procedures like needle aspiration. A female host and a nurse are present to conduct the interview.

\tcblower 

\textbf{Error Type:} Attribute Hijacking \\[0.5em]
\textbf{Generated Query:} \\
Who is the expert being interviewed in the video, what medical condition is discussed, and what are the backgrounds or settings shown during the interview?

\end{tcolorbox}

\vspace{0.5em}

\begin{tcolorbox}[
    colback=gray!5!white,      
    colframe=gray!60!black,    
    title={Example 3: Out-of-Domain Irrelevance}, 
    fonttitle=\bfseries,       
    breakable,                 
    boxrule=0.8pt,            
    arc=3pt,                  
    left=4pt, right=4pt, top=4pt, bottom=4pt 
]

\textbf{Ground Truth:} \\
A news anchor presents a story about a political controversy involving a leaked audio recording. The broadcast displays images of politicians, including President Yoon Suk-yeol, and shows text from a social media post by Lee Jun-seok, who denies being the source of the leak. A press conference is shown where a masked man speaks at a podium in front of the National Assembly seal, addressing the allegations. The report includes an animated graphic depicting two silhouetted figures representing lawmakers from the People Power Party, discussing the situation.

\tcblower 

\textbf{Error Type:} Out-of-Domain Irrelevance \\[0.5em]
\textbf{Generated Query:} \\
What is the main topic of the news report in the video?

\end{tcolorbox}

\vspace{0.5em}

\begin{tcolorbox}[
    colback=gray!5!white,      
    colframe=gray!60!black,    
    title={Example 4: Information Sparsity}, 
    fonttitle=\bfseries,       
    breakable,                 
    boxrule=0.8pt,            
    arc=3pt,                  
    left=4pt, right=4pt, top=4pt, bottom=4pt 
]

\textbf{Ground Truth:} \\
The video is a news report from YTN about a political controversy involving the People Power Party. It features a female anchor introducing the story, followed by on-screen text messages allegedly exchanged between party members discussing the possibility of a candidate's withdrawal. The report includes footage of a press conference with Kim Dong-cheol, the party's floor leader, who denies wrongdoing and claims the matter was handled internally. Other party figures, including Lee Yong-joo and Lee Sang-tae, are shown speaking at events, while opposition leaders like Park Hee-ryeon and Ahn Cheol-soo are also featured. The segment concludes with a reporter providing an update on the situation outside a government building.

\tcblower 

\textbf{Error Type:} Information Sparsity \\[0.5em]
\textbf{Generated Query:} \\
What are the specific details and sequence of events reported in this news segment about the political controversy?

\end{tcolorbox}

\end{document}